\documentclass[runningheads]{llncs}
\usepackage{arydshln}
\usepackage{adjustbox}
\usepackage{multirow}
\usepackage{graphicx}
\usepackage{subcaption}
\usepackage{tabularx}
\usepackage[table,xcdraw]{xcolor}
\newcommand\blfootnote[1]{%
  \begingroup
  \renewcommand\thefootnote{}\footnote{#1}%
  \addtocounter{footnote}{-1}%
  \endgroup
}

\begin{document}
\title{Advances in Human-Robot Handshaking}
%
%
\author{Vignesh Prasad*\inst{1} \orcidID{0000-0002-9729-2454} \and \\
Ruth Stock-Homburg\inst{1} \orcidID{0000-0002-8576-5883} \and \\ 
Jan Peters\inst{1,2} \orcidID{0000-0002-5266-8091}}%
\authorrunning{V. Prasad et al.}
\institute{Technical University of Darmstadt, Germany \and
Max Planck Institute for Intelligent Systems, T\"ubingen, Germany
\email{\{vignesh.prasad,ruth.stock-homburg,jan.peters\}@tu-darmstadt.de}\\
}
\maketitle              

\begin{abstract}
The use of social, anthropomorphic robots to support humans in various industries has been on the rise. During Human-Robot Interaction (HRI), physically interactive non-verbal behaviour is key for more natural interactions. Handshaking is one such natural interaction used commonly in many social contexts. It is one of the first non-verbal interactions which takes place and should, therefore, be part of the repertoire of a social robot. In this paper, we explore the existing state of Human-Robot Handshaking and discuss possible ways forward for such physically interactive behaviours.
\end{abstract}
%
%
%
\blfootnote{* - Corresponding Author}

\section{Introduction}
Handshaking is a commonly and naturally used physical interaction and an important social behaviour between two people \cite{deborah1974handwork} in many social contexts \cite{chaplin2000handshaking,hall1983handshake,stewart2008exploring}. It is one of the most common greetings that is usually the first non-verbal interaction taking place in a social context. Handshaking is, therefore, an important social cue for several reasons. Firstly, it plays an important role in shaping impressions \cite{bernieri2011influence,chaplin2000handshaking,stewart2008exploring}. Moreover, it helps set the tone of any interaction, since the sense of touch can convey distinct emotions \cite{hertenstein2006communicative}. Robot handshaking improves the perception of robots as well by making humans more willing to help them \cite{avelino2018power} leading to better cooperation and coexistence.

Having human-like body movements plays an important role in the acceptance of HRI as well. Thus, having a good handshake can not only widen the expressive abilities of a social robot but also provide a strong first impression for further interactions to take place. 

\begin{figure}
    \centering
    \includegraphics[width=\textwidth]{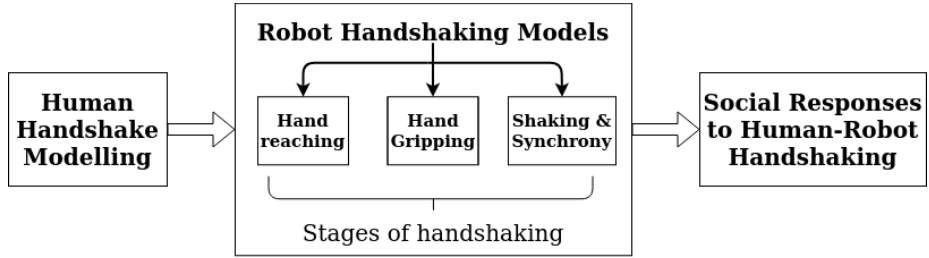}
    \caption{Conceptual Framework for Human-Robot Handshaking}
    \label{fig:framework}
\end{figure}

We propose the following framework in our study, shown in Fig. \ref{fig:framework}. 

\section{Handshake Modelling}
\label{sec:modelling}

A group of researchers from Okayama Prefectural University, Japan modelled human handshaking interactions using motion capture to track participants' joints. Firstly, a transfer function to mimic the requester's reaching motion for the responder was developed \cite{jindai2006development}. This was further developed into a minimum jerk trajectory model which accurately captures the velocity profiles and generates smooth motions \cite{jindai2015handshake,jindai2011development,ota2014handshake,ota2015handshake}. Shaking was modelled as a spring and damper system \cite{jindai2008handshake,yamato2008development}, whose oscillatory motion profile fit that of shaking. A similar spring-damper model is proposed by Dai et al. \cite{dai2019research} to model the elbow stiffness by measuring muscle contractions in the arm using EMG signals. 

A group of researchers from the University of Lorraine on modelled the mutual synchronization (MS in short) between participants during shaking as well as the forces exerted on the palms \cite{melnyk2014analysis,melnyk2019physical,tagne2016measurement}. 
Tagne et al. \cite{tagne2016measurement} investigate the joint motions with IMUs place at each joint. Melnyk and H\'enaff \cite{melnyk2019physical} trends across different gender pairings are analysed as well. Both of these works analyze the influences of social setting, such as greeting, congratulating or sympathising.
Tagne et al. \cite{tagne2016measurement} observe a shorter duration during greeting compared to sympathy and congratulation, which were similar. The grip strength shows contradictory results. Tagne et al. \cite{tagne2016measurement} observe the lowest grip strength in case of sympathy, followed by greeting and then, congratulations. Melnyk and H\'enaff \cite{melnyk2019physical} observe slightly higher grip strength for consolation although not significantly. Regarding gender, male pairs shook for a lesser duration than mixed pairs and female pairs shook the longest. 
No conclusive correlations were found between gender and grip strength, contrary to \cite{chaplin2000handshaking,orefice2016let}.

Knoop et al. \cite{knoop2017handshakiness} studied the contact area, pressure and grasping forces exerted while handshaking. 
A positive correlation was found between contact pressure and grasping force. This was non-linear as the grasping force got higher.

\section{Hand Reaching in Handshaking}
\label{sec:reaching}
Jindai et al. \cite{jindai2015handshake,jindai2011development,jindai2006development} and Ota et al. \cite{ota2014handshake,ota2015handshake} propose two models for reaching. One with a transfer function based on the human hand's trajectory with a lag element and the other is a minimum jerk trajectory model, which fits the velocity profiles and provides smooth trajectories by definition. 

More recently, works model reaching using machine learning. Campbell et al. \cite{campbell2019learning} use imitation learning to learn a joint distribution over the actions of both the human and the robot during handshaking. They execute open-loop trajectories and human adjusts to them in training as their pneumatic robot cannot be kinesthetically taught. During test time, the posterior distribution is inferred from the human's initial motions from which the robot's trajectory is sampled. Their framework estimates the speed of the interaction as well, to match the speed of the human. Christen et al. \cite{christen2019guided} use Deep Reinforcement Learning to learn physical interactions from human-human interactions. They use an imitation reward which helps in learning the intricacies of the interaction. Falahi et al. \cite{falahi2014adaptive} use one-shot imitation learning to kinesthetically teach reaching and shaking behaviours based on gender and familiarity detected using facial recognition. However, it cannot be generalized due to the extremely low sample size. Vinayavekhin et al. \cite{vinayavekhin2017human} model hand reaching with an LSTM trained using skeleton data. They predict the human hand's final pose and devise a simple controller for the robot arm to reach the predicted location. In terms of smoothness, timeliness and efficiency, their method performs better than following the intermediate hand locations. However, it performs worse than using the true final pose due to inaccuracies in the prediction.


\section{Hand Grasping Control}
\label{sec:grasping}
Avelino et al. \cite{avelino2017human} model grasping with different degrees of hand closure. Since position control is used, the force perceived depends on the hand sizes of the participants. They address this in their next work \cite{avelino2018towards}, where participants adjust the robot's fingers until a preferable grasp is reached. This provides a reference for the force sensors and the joint positions, using which grasping behaviours are developed. They do not incorporate any force-feedback, which is discussed below.

Ouchi and Hashimoto \cite{ouchi1997handshake} propose a remote handshaking system, where a handshake is performed while on a call using a custom silicone-rubber based robotic soft hand. The force exerted on the robot hand, measured using a pneumatic force sensor, is relayed to the robot hand at the other end that mimics it. They see that participants barely felt any transmission delay and perceived the partner's existence better during the call. Pedemonte et al. \cite{pedemonte2016design} design a robot hand for handshaking controlled by the force exerted on it. It is sensor-less, with a deformable palm controlling the fingers based on the degree of deformation using variable admittance control. Arns et al. \cite{arns2017design} improve this design with lower gear ratios, impedance control and more powerful actuators to obtain stronger grasping forces and almost instantaneous speeds ($<$ 0.05s). The force exerted by the robot hand depends on the force exerted by the human, leading to a partial synchronisation. Arns et al. test how it feels in comparison with a human hand on a 5-point scale (1-very different, 5-identical). It was perceived well in terms of compliance (3.9/5), force feedback (4/5) and overall haptics (3.7/5).

Vigni et al. \cite{vigni2019role} follow a more closed-loop approach by measuring the force exerted by the robot hand with force-sensitive resistors and control the force exerted by the robot hand which is approximated from the degree of hand closure. They compare three different relationships between the exerted forces of the human and the robot namely linear, constant and combined (constant+linear). The latter two are used with high (strong) and low (weak) constant values. The combined controllers were perceived better than the constant ones. Participants were seen to adjust their force based on the robot's, showing that humans tend to follow the force exerted on their hand. The stronger variants of the constant and combined controllers were perceived as more confident/extroverted. 

\section{Shaking and Synchronisation}
\label{sec:shaking}
In this section, we describe works that that model the shaking phase. They mainly do so by aiming to achieve synchronous motions with the interaction partner while reducing interaction forces. The works can be broadly divided into the following three categories: Central Pattern Generator Models, Harmonic Oscillator Models and Miscellaneous Models.


\subsection{Central Pattern Generators} 
Central Pattern Generators (CPGs) \cite{hooper2001central} are biologically inspired neuronal circuits, that generate rhythmic outputs from non-rhythmic inputs. Kasuga and Hashimoto \cite{kasuga2005human} model the shoulder and elbow motions of a robot using the exerted torque on the joints as input with a CPG which, however, doesn't adapt to the human. For better synchronization, some works adapt the CPG to learn the frequency of the shaking motions. This is achieved by either incorporating a learning framework into the CPG \cite{artem2013physical,jouaiti2018hebbian,melnyk2016bio} or by parametrizing the CPG and learning the parameters on the fly \cite{papageorgiou2015kinematic,sato2007synchronization}.


\subsection{Harmonic Oscillator Models} 
Harmonic oscillator models either mimic harmonic systems like spring-damper systems\cite{dai2019variable,mura2020role,yamato2008development} or follow simple sinusoidal motions \cite{beaudoin2019haptic,wang2009hmm,wang2008modelling,zeng2012human}. Chua et al. \cite{chua2010human} propose a hybrid model that uses both, a spring-damper model to update impedance parameters and a simple sinusoidal trajectory to generate reference trajectories. Similarly other works use impedance control to model the stiffness \cite{beaudoin2019haptic,dai2019variable} and for better synchronization, some  estimate the impedance parameters in an online fashion by using an EKF \cite{mura2020role}, a HMM \cite{wang2009hmm} or least-squares \cite{wang2008modelling}. 


\subsection{Miscellaneous Models} Karniel et al. \cite{karniel2010turing} and Nisky et al. \cite{nisky2012three} design an experimental framework and metric for testing the human-likeness of shaking motions on a 1D haptic stylus. Avraham et al. \cite{avraham2012toward} test 3 models with this. The first is a tit-for-tat model that passively records the joint motions and replays it. The second is a biological model simulating muscle generated motions to achieve low interaction forces. The final is a simple linear regression model. The tit-for-tat and linear regression models fare much better than the biologically inspired model.

Pedemonte et al. \cite{pedemonte2017haptic} introduce a remote handshaking mechanism using their previously developed hand. They develop a vertical rail mechanism that the hand is mounted on and is passively controlled. This shaking motion along with the force exerted on the hand is relayed to the partner's hand and rail mechanism. This allows realistic haptic interaction to take place remotely where the participants can adequately perceive each other's motions and forces.

\section{Social Responses to Human-Robot Handshaking}
\label{sec:social}
Before we talk about the various social responses, we would like to discuss the differences in metrics and criteria used for understanding the way different methods evaluate their studies. One common metric used is the bipolar scale (7-point or 5-point scale) where one end conveys a negative perception of the parameter, and the other end conveys a positive perception. Another popular method is the Bradley-Terry model \cite{bradley1952rank}, which is a probabilistic model specifically used for understanding paired comparisons among a set of different methods. However, these are general metrics used for statistical analysis. To this end Karniel et al. \cite{karniel2010turing} devise a custom metric for comparing the human-likeness of different human-robot handshaking methods in a Turing test-like setting, called a Model Human Likeness Grade (MHLG). This is based on the perceived probability of a method being human-like by a participant. 

Additionally, the use of many different types of robotic interfaces makes it difficult to generalize the comparison of results across different works. Some works, use a simple gripper like interface \cite{bevan2015shaking,falahi2014adaptive,jouaiti2018hebbian,sato2007synchronization}, some use a rod-like end-effector \cite{avraham2012toward,giannopoulos2011comparison,karniel2010turing,nisky2012three,papageorgiou2015kinematic,wang2009hmm,wang2011handshake}, and some use a human-hand like interface that is either actively controlled \cite{avelino2018power,avelino2018towards,ammi2015haptic,christen2019guided,mura2020role,tsalamlal2015affective,vigni2019role}, passively controlled \cite{arns2017design,beaudoin2019haptic,ouchi1997handshake,pedemonte2016design,pedemonte2017haptic} or not controlled at all \cite{campbell2019learning,dai2019variable,jindai2015handshake,jindai2012small,jindai2007development,jindai2008handshake,jindai2010small,jindai2011development,jindai2006development,kasuga2005human,knoop2017handshakiness,melnyk2016bio,nakanishi2014remote,orefice2018pressure,ota2014handshake,ota2015handshake,stock2020evaluation,vanello2010neural,vinayavekhin2017human,yamato2008development} 

Such a variety in the usage of different evaluation criteria, metrics and especially robotic interfaces makes it is difficult to converge on a common benchmark on common parameters to evaluate different human-robot handshaking methods. Therefore, we categorize the different works evaluating human-robot handshaking based on the factors they evaluate or the goal of their experiments. These can roughly be divided into the categories as shown below.

\subsection{Influence of External Factors}
Ammi et al. \cite{ammi2015haptic} and  Tsalamlal et al. \cite{tsalamlal2015affective} explored combinations of visual and haptic behaviours. Among visual expressions, "happy" was rated higher than "neutral" one, the least being "sad". Significantly higher arousal and dominance were seen for strong handshakes and higher valence for soft ones. Higher arousal and dominance was also seen with strong handshakes in a visuo-haptic case as compared to a visual-only case. Another framework studying the effect of visuo-haptic stimuli was proposed by Vanello et al. \cite{vanello2010neural}. They develop a sensor glove to track the participants' hand motions and contact pressure and have a screen on which visuals (faces of humans and robots) are shown. Participants' feedback is analysed using fMRI activity. Nothing can be concluded from their results as only three participants took part in their study. 

Jindai et al. \cite{jindai2007development}, found that a delay of 0.1 seconds between the voice and handshake motion of the robot was acceptable. Jindai et al.  \cite{jindai2011development}, they further saw that participants preferred when the gaze shifted steadily from the hand while reaching to the face after contact. Ota et al. \cite{ota2014handshake,ota2015handshake} found the response to a handshake to be preferable with a delay of 0.2s to 0.4s.

Nakanishi et al. \cite{nakanishi2014remote} equip a robotic soft hand on a video screen showing a remote presenter in a telepresence scenario. Interactions were better perceived when the presenter's hand was not visible on the screen. They further saw that when participants controlled a second robot hand placed with the presenter, feelings of closeness and physically shaking hands were rated higher when both the presenter's hand and the robot hand were not in the frame. They argue that the hand's visibility cancels the feeling of synchronization, which some subjects reported was due to seeing two hands for the same interaction.

\subsection{Influence of Handshaking on Robot Perception}

Avelino et al. \cite{avelino2018power} see how handshaking affects the willingness to help a robot when it has to perform a navigation task. Participants who shook hands with the robot found it to be more warm and likeable and were more willing to help the robot. However, they argue that a human-like robot handshake would lead to participants not anticipating the robot getting stuck in a simple navigational task due to a mismatch between the expected skill and the actual behaviour. Bevan and Fraser \cite{bevan2015shaking} study the effect of handshaking on negotiations between participants, where one participant interacts remotely via a Nao robot. Handshaking improved mutual cooperation, however, haptic feedback for the telepresent negotiator had no significant impact. It did not affect the perceived trustworthiness, which they argue is possibly due to the childlike nature of the Nao robot.

\subsection{Using Handshaking for Differentiation}
Garg et al. \cite{garg2017classifying} classify people's personality as introverts and extroverts using statistics of accelerations, Euler angles and polar orientations when shaking hands with a robot hand. The features are ranked based on  Mutual Information followed by a K Nearest Neighbours classification, achieving a 75\% accuracy. Orefice et al. \cite{orefice2016let} similarly look at distinctions in personality as well as gender. They found that male-male pairs applied more pressure than male-female pairs. Female-female pairs had a longer duration and lower frequency. Regarding personality, they found that introverts shook at higher speeds while extroverts exerted more pressure. Using these features, they predict the human's gender and personality during a human-robot handshake. 

They further perform a longitudinal study \cite{orefice2018pressure} that looks at how pressure variations while shaking hands with a Pepper robot reflects the participants' immediate mood. A consistency was seen when shaking hands with a human subject or with Pepper, which was unexpected as interacting with Pepper might not be as human-like. The only significant differences between different positive moods were with "Calm" and "Cheerful" moods, with less pressure observed in a "Calm" mood. For negative moods, a "Bored" mood had lower pressure than "Excited" or "Tense" moods, both of which had more arousal. In general, lower pressures were found with moods with lower arousal. 

\subsection{Human-likeness Evaluation}
Giannopoulos et al. \cite{giannopoulos2011comparison} and Wang et al. \cite{wang2011handshake}  compare the human likeness of their previous handshaking models (a basic one\cite{wang2008modelling} and an interactive one\cite{wang2009hmm}) with a human operating the robot. Both studies perform their experiment with participants wearing noise-cancelling headphones playing music and ambient conversations in a cocktail bar scenario. Giannopoulos et al. \cite{giannopoulos2011comparison} blindfold the participants and Wang et al. \cite{wang2011handshake} make the participants wear a VR headset with a human model rendered for the robot. The human-operated handshake was rated the highest (6.8/10 in both), followed by the interactive handshake (5.9/10 in \cite{giannopoulos2011comparison}, 5.3/10 in \cite{wang2008modelling}). The basic handshake was rated the lowest (3.3/10 in \cite{giannopoulos2011comparison}, 3.0/10 in \cite{wang2008modelling}). The interactive handshake was close to the human-operated one, but both were far from the maximum human-likeness (10/10), possibly due to the rod-like end effector.

Stock-Homburg et al. \cite{stock2020evaluation} test if an android robot's hand, made with soft silicone skin and a heated palm, can pass as a human hand. Participants were blindfolded and shook hands with a human and the robot twice in random order. Majority of them (11/15) correctly guessed the first hand they interacted with, which some said was from the mechanical feel of the robot hand. By the fourth handshake, all guessed correctly. Participants only had a static interaction. Testing handshake behaviours instead, could yield better insights.

\section{Discussion and Conclusion}

Overall, we discussed various works looking into human-robot handshaking. Due to differences in hardware and metrics, it is difficult to come up with a common benchmark to evaluate these studies. However, some qualitative conclusions can be drawn. In general, an element of synchronization is present. This can be measured well in the shaking stage where low interaction forces can be an indication of synchronization. In reality, there is a leader-follower situation which arises, which could perhaps reflect on various personal attributes of the people shaking hands. From the perspective of a social robot, contextual cues would be effective in having a better impact from the handshake. This requires further research in other fields like emotion recognition, estimating intent, personality etc. For a more human-like perception, each aspect of the movement at different stages needs to be human-like since we are still far from having robotic interfaces that not only look human-like but also feel human-like, as most still have a mechanical feel. This combined with a smooth integration of the different phases of handshaking is also important since delays in switching between the different stages could possibly not be well perceived.  

One thing to keep in mind is physical interactions vary over different cultures, age groups, geographic locations. Depending on the context too, different interactions would be more prevalent, like hugging or patting for higher intimacy or bumping fists or giving high fives in a friendly scenario. Moreover, due to the Covid-19 pandemic, there are increasing restrictions and precautions regarding limiting physical contact which has led to alternative interactions, like shaking/tapping feet, touching elbows/forearms, remote high fives and so on. However, the importance of handshakes in business and formal settings is a good motivation for continuing to develop human-like handshaking behaviours. Additionally, learning different physically interactive behaviours would help improve the perception of a social robot, which is a good direction for future work.

\section{Acknowledgements}
The authors thank the Interdisciplinary Research Forum (Forum Interdisziplin\"are Forschung) at the Technical University of Darmstadt and the Association of Supporters of Market-Oriented Management, Marketing, and Human Resource Management (F\"orderverein f\"ur Marktorientierte Unternehmensf\"uhrung, Marketing und Personalmanagement e.V.) for funding this work.

%
%
%
\bibliographystyle{splncs04}
\bibliography{references}

\end{document}